\documentclass[11pt,a4paper]{article}
\usepackage[hyperref]{acl2018}
\usepackage{times}
\usepackage{latexsym}
\usepackage{graphicx}
\usepackage{url}
\usepackage{amsfonts}
\usepackage{amsmath}

\usepackage{url}

\aclfinalcopy 
\def\aclpaperid{1267} 

\newcommand{\ignore}[1]{}

\newcommand{\sm}[1]{\textcolor{red}{\bf\small [#1 --SM]}}

\newcommand{\excont}[1]{{\mbox{\hspace{0.2in}\bf\it #1 }}}

\newcommand{\Sref}[1]{\S\ref{#1}}
\newcommand{\fref}[1]{Figure~\ref{#1}}

\newcommand{\tref}[1]{Table~\ref{#1}}

\newcommand{\possessivecite}[1]{\citeauthor{#1}'s (\citeyear{#1})}
\title{Style Transfer Through Back-Translation}

\author{Shrimai Prabhumoye, Yulia Tsvetkov, 
    Ruslan Salakhutdinov, Alan W Black\\
  Carnegie Mellon University, Pittsburgh, PA, USA\\
  {\tt \{sprabhum,ytsvetko,rsalakhu,awb\}@cs.cmu.edu}
  }

\date{}

\begin{document}
\maketitle
\begin{abstract}
Style transfer is the task of rephrasing the text to contain specific stylistic properties without changing the intent or affect within the context. 
This paper introduces a new method for automatic style transfer. 
We first learn a latent representation of the input sentence which is grounded in a language translation model in order to better preserve the meaning of the sentence while reducing stylistic properties.
Then adversarial generation techniques are used to make the output match the desired style. 
We evaluate this technique on three different style transformations: sentiment, gender and political slant.  
Compared to two state-of-the-art style transfer modeling techniques we show improvements both in automatic evaluation of style transfer and in manual evaluation of meaning preservation and fluency. 

\end{abstract}

\section{Introduction}

Intelligent, situation-aware applications must produce
naturalistic outputs, lexicalizing the same meaning
differently, depending upon the environment. This is particularly relevant for language generation tasks
such as machine translation \cite{sutskever2014sequence,bahdanau2015neural},
caption generation \cite{karpathy2015deep,xu2015show},
and natural language generation \citep{wen2017network,kiddon2016globally}.
In conversational agents \cite{ritter2011data,sordoni2015neural,vinyals2015,li2016persona},
for example, modulating the politeness style, to sound natural depending upon a situation: at a party with friends ``Shut up! the video is starting!'', or in a professional setting ``Please be quiet, the video will begin shortly.''. 

\begin{figure*}[t]
\centering
\includegraphics[scale=0.35]{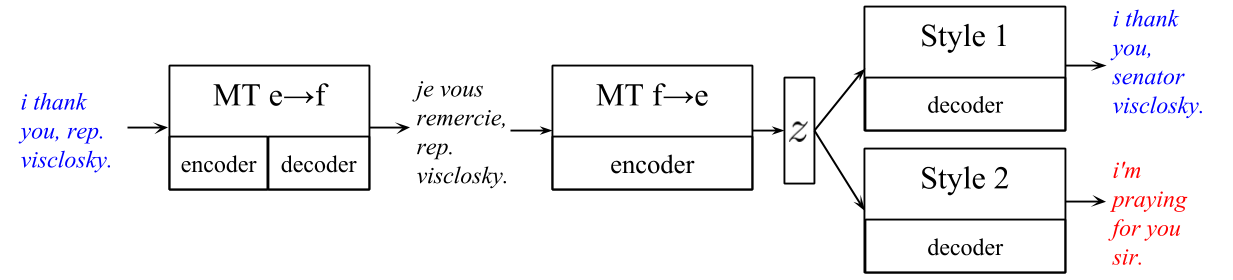}
\caption{Style transfer pipeline: to rephrase a sentence and reduce its stylistic characteristics, the sentence is back-translated. Then, separate style-specific generators are used for style transfer.}
\label{overview}
\vspace{-1em}
\end{figure*}

These goals have motivated a considerable amount 
of recent research efforts focused at ``controlled'' 
language generation---aiming at separating the semantic content 
of \emph{what} is said from the stylistic dimensions of \emph{how} 
it is said. These include approaches
relying on heuristic substitutions, deletions, and insertions to
modulate demographic properties of a writer \citep{reddy2016obfuscating},
integrating stylistic and demographic speaker traits
in statistical machine translation \cite{rabinovich2016personalized,niu2017study},  
and deep generative models controlling for a particular stylistic
aspect, e.g., politeness \cite{sennrich2016controlling}, 
sentiment, or tense  \cite{hu2017toward,shen2017style}. 
The latter approaches to style transfer, while more 
powerful and flexible than heuristic methods, have yet to show that in addition to transferring style they effectively preserve meaning of input sentences. 

\ignore{
This paper introduces a novel approach to transferring style 
of a sentence while preserving its meaning.  
Parallel text with the same meaning but different styles is rarely available. 
Thus, we are deliberately concentrating on style transfer with non-parallel (monolingual) corpora.  We consider other style transfer techniques that depend on parallel training data (either from paraphrase copora or from translated lexicons) to be a different problem. \sm{Do we need this para?}
}

This paper introduces a novel approach to transferring style 
of a sentence while better preserving its meaning.  
We hypothesize---relying on the study of \citet{rabinovich2016personalized} 
who showed that author characteristics are significantly obfuscated
by both manual and automatic machine translation---that grounding in back-translation 
is a plausible approach to rephrase a sentence while reducing its  
stylistic properties.  
We thus first use back-translation to rephrase the sentence and reduce the effect of the original style; then, we generate from the latent representation, using separate style-specific generators controlling for style (\Sref{sec:methodology}). 

We focus on transferring author attributes: (1)~gender and (2)~political slant, and (3)~on sentiment modification. 
The second task is novel: 
given a sentence by an author with a particular political leaning, 
rephrase the sentence to preserve its meaning but to confound 
classifiers of political slant (\Sref{sec:tasks}). 
The task of sentiment modification enables us to compare our approach 
with state-of-the-art models \cite{hu2017toward,shen2017style}.

Style transfer is evaluated using style classifiers trained on held-out data. 
Our back-translation style transfer model outperforms the state-of-the-art baselines \cite{shen2017style,hu2017toward} on the tasks of political slant and sentiment modification; 12\% absolute improvement was attained for political slant transfer, and up to 7\% absolute improvement in modification of sentiment (\Sref{sec:results}).
Meaning preservation was evaluated manually, using A/B testing (\Sref{sec:experiments}). 
Our approach performs better than the baseline on the task of transferring gender and political slant. 
Finally, we evaluate the fluency of the generated sentences using human evaluation and our model outperforms the baseline in all experiments for fluency.

\begin{figure*}[t]
\centering
\includegraphics[scale=0.35]{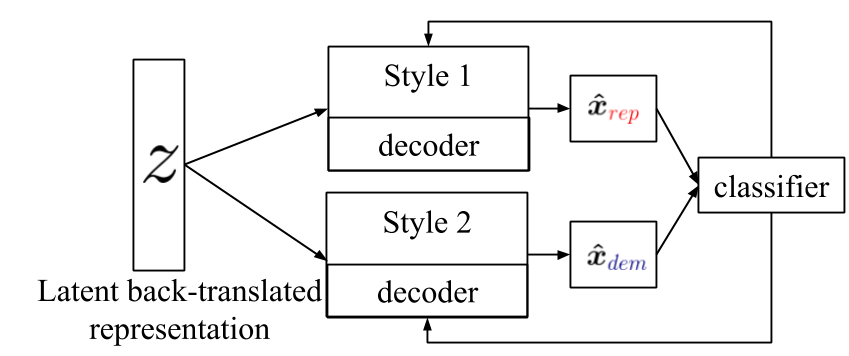}
\caption{The latent representation from back-translation and the style classifier feedback are used to guide the style-specific generators.}
\label{latent}
\vspace{-1em}
\end{figure*}

The main contribution of this work is a new approach to style transfer that outperforms state-of-the-art baselines in both 
the quality of input--output correspondence (meaning preservation and fluency), 
and the accuracy of style transfer. The secondary contribution is a new task that we propose to evaluate style transfer: transferring political slant.

\section{Methodology}
\label{sec:methodology}

Given two datasets $\boldsymbol{X}_1 = \{\boldsymbol{x}_{1}^{(1)},\ldots,\boldsymbol{x}_{1}^{(n)}\}$ and $\boldsymbol{X}_2 = \{\boldsymbol{x}_{2}^{(1)},\ldots,\boldsymbol{x}_{2}^{(n)}\}$ which represent two different styles $\boldsymbol{s}_1$ and $\boldsymbol{s}_2$, respectively, our task is to generate sentences of the desired style while preserving the meaning of the input sentence. Specifically, we generate samples of dataset $\boldsymbol{X}_1$ such that they belong to style $\boldsymbol{s}_2$ and samples of $\boldsymbol{X}_2$ such that they belong to style $\boldsymbol{s}_1$. We denote the output of dataset $\boldsymbol{X}_1$ transfered to style $\boldsymbol{s}_2$ as $\hat{\boldsymbol{X}_1} = \{\hat{\boldsymbol{x}}_{2}^{(1)},\ldots,\hat{\boldsymbol{x}}_{2}^{(n)}\}$ and the output of dataset $\boldsymbol{X}_2$ transferred to style $\boldsymbol{s}_1$ as $\hat{\boldsymbol{X}_2} = \{\hat{\boldsymbol{x}}_{1}^{(1)},\ldots,\hat{\boldsymbol{x}}_{1}^{(n)}\}.$

\citet{hu2017toward} and \citet{shen2017style} introduced state-of-the-art style transfer models that use variational auto-encoders \cite[VAEs]{kingma2013auto} and cross-aligned auto-encoders, respectively, to model a latent content variable $\boldsymbol{z}$. The latent content variable $\boldsymbol{z}$ is a code which is not observed. The generative model conditions on this code during the generation process. Our aim is to design a latent code~$\boldsymbol{z}$ which (1) represents the meaning of the input sentence grounded in back-translation and (2) weakens the style attributes of author's traits. To model the former, we use neural machine translation. Prior work has shown that the process of translating a sentence from a source language to a target language retains the meaning of the sentence but does not preserve the stylistic features related to the author's traits \cite{rabinovich2016personalized}. We hypothesize that a latent code~$\boldsymbol{z}$ obtained through back-translation will normalize the sentence and devoid it from style attributes specific to author's traits.

Figure \ref{overview} shows the overview of the proposed method. In our framework, we first train a machine translation model from source language~$e$ to a target language~$f$. We also train a back-translation model from~$f$ to~$e$. 
Let us assume our styles $\boldsymbol{s}_1$ and $\boldsymbol{s}_2$ correspond to \textsc{democratic} and \textsc{republican} style,  respectively. 
In Figure \ref{overview}, the input sentence \textit{i thank you, rep. visclosky.} is labeled as \textsc{democratic}. We translate the sentence using the $e \rightarrow f$ machine translation model and generate the parallel sentence in the target language $f$: \textit{je vous remercie, rep. visclosky.} Using the fixed encoder of the $f \rightarrow e$ machine translation model, we encode this sentence in language $f$. The hidden representation created by this encoder of the back-translation model is used as $\boldsymbol{z}$. We condition our generative models on this $\boldsymbol{z}$. We then train two separate decoders for each style $s_1$ and $s_2$ to generate samples in these respective styles in source language $e$. Hence the sentence could be translated to the \textsc{republican} style using the decoder for $\boldsymbol{s}_2$. For example, the sentence \textit{i'm praying for you sir.} is the \textsc{republican} version of the input sentence and \textit{i thank you, senator visclosky.} is the more \textsc{democratic} version of it.

Note that in this setting, the machine translation and the encoder of the back-translation model remain fixed. They are not dependent on the data we use across different tasks. This facilitates re-usability and spares the need of learning separate models to generate $\boldsymbol{z}$ for a new style data. 

\subsection{Meaning-Grounded Representation}

In this section we describe how we learn the latent content variable $\boldsymbol{z}$ using back-translation. 
The $e \rightarrow f$ machine translation and $f \rightarrow e$ back-translation models are trained using a sequence-to-sequence framework \cite{sutskever2014sequence,bahdanau2015neural} with style-agnostic corpus. 
The style-specific sentence \textit{i thank you, rep. visclosky.} in source language $e$ is translated to the target language $f$ to get \textit{je vous remercie, rep. visclosky.} The individual tokens of this sentence are then encoded using the encoder of the $f \rightarrow e$ back-translation model. The learned hidden representation is $\boldsymbol{z}$. 

Formally, let $\boldsymbol{\theta}_{E}$ represent the parameters of the encoder of $f \rightarrow e$ translation system. Then $\boldsymbol{z}$ is given by:
\begin{equation}
\boldsymbol{z} = Encoder(\boldsymbol{x}_{f}; \boldsymbol{\theta}_{E})
\label{z_rep}
\end{equation}
where, $\boldsymbol{x}_{f}$ is the sentence $\boldsymbol{x}$ in language $f$. Specifically, $\boldsymbol{x}_{f}$ is the output of $e \rightarrow f$ translation system when $\boldsymbol{x}_{e}$ is given as input. Since $\boldsymbol{z}$ is derived from a non-style specific process, this Encoder is not style specific.

\subsection{Style-Specific Generation}
\label{sec:style-specific-generator}

Figure \ref{latent} shows the architecture of the generative model for generating different styles. Using the encoder embedding $\boldsymbol{z}$, we train multiple decoders for each style. The sentence generated by a decoder is passed through the classifier. The loss of the classifier for the generated sentence is used as feedback to guide the decoder for the generation process. The target attribute of the classifier is determined by the decoder from which the output is generated. For example, in the case of \textsc{democratic} decoder, the target attribute is \textsc{democratic} and for the \textsc{republican} decoder the target is \textsc{republican}. 

\ignore{In section \ref{sec:classifier} we describe the learning process of the classifier and in section \ref{sec:generator} we describe the learning of the generative model and how the trained classifier is used to guide the generator.}

\subsubsection{Style Classifiers}
\label{sec:classifier}

We train a convolutional neural network (CNN) classifier to accurately predict the given style. 
We also use it to evaluate the error in the generated samples for the desired style. We train the classifier in a supervised manner. 
The classifier accepts either discrete 
or continuous tokens as inputs. 
This is done such that the generator output can be used as input to the classifier. 
We need labeled examples to train the classifier 
such that each instance in the dataset $\boldsymbol{X}$ should 
have a label in the set $\boldsymbol{s} = \{\boldsymbol{s}_1, \boldsymbol{s}_2\}$. 
Let $\boldsymbol{\theta}_{C}$ denote the parameters of the classifier. 
The objective to train the classifier is given by:
\begin{equation}
\mathcal{L}_{class}(\boldsymbol{\theta}_{C}) = \mathbb{E}_{\boldsymbol{X}} [\log q_{C} (\boldsymbol{s}|\boldsymbol{x})].
\label{class_loss}
\end{equation}
To improve the accuracy of the classifier, we augment classifier's inputs with style-specific lexicons.  
We concatenate binary style indicators to each input word embedding in the classifier. The indicators are set to 1 if the input word is present in a style-specific lexicon; otherwise they are set to 0. 
Style lexicons are extracted using the log-odds ratio informative Dirichlet prior \cite{monroe_colaresi_quinn}, a method that identifies words that are statistically overrepresented in each of the categories. 

\ignore{: \begin{eqnarray*}
\hat{\delta}_{w}^{(i-j)} = \log(\frac{y_{w}^{i} + \alpha_{w}}{n^{i} + \alpha_{0} - (y_{w}^{i} + \alpha_{w})}) - \\
\log(\frac{y_{w}^{j} + \alpha_{w}}{n^{j} + \alpha_{0} - (y_{w}^{j} + \alpha_{w})}),
\end{eqnarray*}
where, $w$ is the word in two corpora $i$ and $j$ and $\hat{\delta}_{w}^{(i-j)} $ is the log odds ratio for $w$, $n^{i}$ and $n^{j}$ are the sizes of the corpora $i$ and $j$ respectively. Similarly, $y_{w}^{i}$ and $y_{w}^{j}$ are the counts of the word $w$ in corpora $i$ and $j$ respectively, $\alpha_0$ is the size of the background corpus, and $\alpha_{w}$ is the count of the word $w$ in the background corpus.} 

\subsubsection{Generator Learning}
\label{sec:generator}

We use a bidirectional LSTM to build our decoders which generate the sequence of tokens $\hat{\boldsymbol{x}} = \{x_1, \cdots x_T\}$. The sequence $\hat{\boldsymbol{x}}$ is conditioned on the latent code $\boldsymbol{z}$ (in our case, on the machine translation model). In this work we use a corpus translated to French by the machine translation system as the input to the encoder of the back-translation model. The same encoder is used to encode sentences of both styles. The representation created by this encoder is given by Eq \ref{z_rep}. Samples are generated as follows:
\begin{eqnarray}
\hat{\boldsymbol{x}} \sim \boldsymbol{z} &=& p(\hat{\boldsymbol{x}} | \boldsymbol{z}) \\
&=& \prod_{t} p(\hat{x}_{t}| \hat{\boldsymbol{x}}^{<t}, \boldsymbol{z})
\end{eqnarray}
where, $\hat{\boldsymbol{x}}^{<t}$ are the tokens generated before $\hat{x}_t$.

Tokens are discrete and non-differentiable. This makes it difficult to use a classifier, as the generation process samples discrete tokens from the multinomial distribution parametrized using softmax function at each time step $t$.
This non-differentiability, in turn, breaks down gradient propagation from the discriminators to the generator. Instead, following \citet{hu2017toward} we use a continuous approximation based on softmax, along with the temperature parameter which anneals the softmax to the discrete case as training proceeds.
To create a continuous representation of the output of the generative model which will be given as an input to the classifier, we use:

\begin{center}
\begin{minipage}{0cm}
\begin{tabbing}
$\hat{x}_{t}  \sim  \text{softmax}(\boldsymbol{o}_{t}/\tau),$
\end{tabbing}
\end{minipage}
\end{center}
where, $\boldsymbol{o}_t$ is the output of the generator and $\tau$ is the temperature which decreases as the training proceeds. Let $\boldsymbol{\theta}_{G}$ denote the parameters of the generators. Then the reconstruction loss is calculated using the cross entropy function, given by:
\begin{equation}
\mathcal{L}_{recon}(\boldsymbol{\theta}_{G}; \boldsymbol{x}) = \mathbb{E}_{q_{E}(\boldsymbol{z}|\boldsymbol{x})} [\log p_{gen}(\boldsymbol{x}|\boldsymbol{z})] \\
\label{recon_loss}
\end{equation}
Here, the back-translation encoder $E$ creates the latent code $\boldsymbol{z}$ by:
\begin{equation}
\boldsymbol{z} = E(\boldsymbol{x}) = q_{E}(\boldsymbol{z}|\boldsymbol{x})
\end{equation}

The generative loss $\mathcal{L}_{gen}$ is then given by:
\begin{eqnarray}
\text{min}_{\theta_{gen}} \mathcal{L}_{gen} &=& \mathcal{L}_{recon} + \lambda_{c}\mathcal{L}_{class}
\end{eqnarray}
where $\mathcal{L}_{recon}$ is given by Eq. (\ref{recon_loss}), $\mathcal{L}_{class}$ is given by Eq (\ref{class_loss}) and $\lambda_{c}$ is a balancing parameter. 

We also use global attention of \cite{DBLP:journals/corr/LuongPM15} to aid our generators. At each time step $t$ of the generation process, we infer a variable length alignment vector $\boldsymbol{a}_t$:
\begin{equation}
\boldsymbol{a}_t = \frac{\text{exp}(\text{score}(\boldsymbol{h}_t, \bar{\boldsymbol{h}}_s))}{\sum_{s^{'}}\text{exp}(\text{score}(\boldsymbol{h}_t, \bar{\boldsymbol{h}}_{s^{'}})}
\end{equation}
\begin{equation}
\text{score}(\boldsymbol{h}_t, \bar{\boldsymbol{h}}_{s}) = \textbf{dot}(\boldsymbol{h}_{t}^{T}, \bar{\boldsymbol{h}}_s),
\end{equation}

where $\boldsymbol{h}_t$ is the current target state and $\bar{\boldsymbol{h}}_{s}$ are all source states.
While generating sentences, we use the attention vector to replace unknown characters (\textsc{unk}) using the copy mechanism in \cite{see2017get}.

\section{Style Transfer Tasks}
\label{sec:tasks}
Much work in computational social science has shown that 
people's personal and demographic
characteristics---either publicly observable 
(e.g., age, gender) or private (e.g., religion, political 
affiliation)---are revealed in their linguistic choices \cite{Nguyen2016}. 
There are practical scenarios, however, when these attributes 
need to be modulated or obfuscated. For example, some users may 
wish to preserve their anonymity online, for personal security 
concerns \cite{jardine2016tor}, or to reduce stereotype threat \cite{Spencer1999}. 
Modulating authors' attributes while preserving 
meaning of sentences can also help generate demographically-balanced 
training data for a variety of downstream applications. 

Moreover, prior work has shown that the quality of 
language identification and POS tagging degrades significantly 
on African American Vernacular English \citep{blodgett2016aae, jorgensen2015challenges}
; YouTube's automatic captions have higher error rates for women and speakers from Scotland~\citep{rudinger2017social}. 
Synthesizing balanced training data---using style transfer 
techniques---is a plausible way to alleviate bias present in existing 
NLP technologies.   

We thus focus on two tasks that have practical and social-good 
applications, and also accurate style classifiers. 
To position our method with respect to prior work, we employ a 
third task of sentiment transfer, which was used in 
two state-of-the-art approaches to style transfer \citep{hu2017toward,shen2017style}. 
We describe the three tasks and associated dataset statistics below.   
The methodology that we advocate is general and can 
be applied to other styles, for transferring various social categories, 
types of bias, and in multi-class settings. 

\paragraph{Gender.} 
In sociolinguistics, gender is known to be one of the most 
important social categories driving language choice \cite{eckert2003language,lakoff2004language,coates2015women}.   
\citet{reddy2016obfuscating} proposed a heuristic-based method 
to obfuscate gender of a writer. This method uses statistical 
association measures to identify gender-salient words and 
substitute them with synonyms typically of the opposite 
gender. This simple approach produces highly fluent, 
meaning-preserving sentences, but does not allow for  
more general rephrasing of sentence beyond single-word 
substitutions. In our work, we adopt this task of 
transferring the author's gender and adapt it to our 
experimental settings. 

We used \possessivecite{reddy2016obfuscating} 
dataset of reviews from Yelp annotated for two genders 
corresponding to markers of sex.\footnote{We note that gender may be considered along a spectrum~\cite{eckert2003language}, but use gender as a binary variable due to the absence of corpora with continuous-valued gender annotations.} 
We split the reviews to sentences, preserving the original 
gender labels. To keep only sentences that are strongly indicative 
of a gender, we then filtered out gender-neutral sentences (e.g., \emph{thank you}) and 
sentences whose likelihood to be written by authors of one gender is lower than 0.7.\footnote{We did not experiment with other threshold values. } 

\paragraph{Political slant.} 

Our second dataset is comprised of top-level comments on Facebook posts from all 412 current members of the United States Senate and House who have public Facebook pages \cite{rtgender}.\footnote{The posts and comments are all public; however, to protect the identity of Facebook users in this dataset \citet{rtgender} have removed all identifying user information as well as Facebook-internal information such as User IDs and Post IDs, replacing these with randomized ID numbers.} Only top-level comments that directly respond to the post are included. Every comment to a Congressperson is labeled with the Congressperson's party affiliation: democratic or republican. 
Topic and sentiment in these comments reveal commenter's political slant. For example, \textit{defund them all,
especially when it comes to the illegal immigrants .} and 
\textit{thank u james, praying for all the work u do .}
are republican,  whereas 
\textit{on behalf of the hard-working nh public school teachers- thank you !} and \textit{we need more strong voices like yours fighting for gun control .} 
represent examples of democratic sentences.
Our task is to preserve intent of the commenter (e.g., to thank their representative), but to modify their observable political affiliation, as in the example in \fref{overview}.  
We preprocessed and filtered the comments similarly to the gender-annotated corpus above. 

\paragraph{Sentiment.}
To compare our work with the state-of-the-art approaches of style transfer for non-parallel corpus we perform sentiment transfer, replicating the models and experimental setups of \citet{hu2017toward} and \citet{shen2017style}. Given a positive Yelp review, a style transfer model will generate a similar review but with an opposite sentiment. 
We used \possessivecite{shen2017style} corpus of reviews from Yelp. 
They have followed the standard practice of labeling the reviews with rating of higher than three as positive and less than three as negative. They have also split the reviews to sentences and assumed that the sentence has the same sentiment as the review.

\paragraph{Dataset statistics.}
We summarize below corpora statistics for the three tasks: transferring gender, political slant, and sentiment. The dataset for sentiment modification task was used as described in \cite{shen2017style}. We split Yelp and Facebook corpora into four disjoint  parts each: (1) a training corpus for training a style classifier (\emph{class}); (2) a training corpus (\emph{train}) used for training the style-specific generative model described in \Sref{sec:style-specific-generator};  (3) development and (4) test sets. We have removed from training corpora \emph{class} and \emph{train} all sentences that overlap with development and test corpora. Corpora sizes are shown in \tref{tab:task-corpora-sizes}.
\begin{table}[t]
\centering
\begin{tabular}{l | c | c | c | c}
\hline
Style & \emph{class} & \emph{train} & \emph{dev} & \emph{test} \\
\hline
gender & 2.57M & 2.67M  & 4.5K & 535K\\
political &  80K & 540K & 4K  &  56K \\
sentiment &  2M & 444K & 63.5K  & 127K  \\
\hline    
\end{tabular}
\caption{Sentence count in style-specific corpora.}
\label{tab:task-corpora-sizes}
\end{table}

Table \ref{tab:vocab_size} shows the approximate vocabulary sizes used for each dataset. The vocabulary is the same for both the styles in each experiment.
\begin{table}[h]
\centering
\begin{tabular}{l | c | c | c}
\hline
Style & gender & political & sentiment \\
\hline
Vocabulary & 20K & 20K  & 10K \\
\hline    
\end{tabular}
\caption{Vocabulary sizes of the datasets.}
\label{tab:vocab_size}
\end{table}

Table \ref{tab:data_stats} summarizes sentence statistics. 
All the sentences have maximum length of 50 tokens. 
\begin{table}[h]
\centering
\begin{tabular}{l | c | c }
\hline
Style & Avg. Length & \%data \\
\hline
male & 18.08 & 50.00  \\
female & 18.21 & 50.00  \\
republican & 16.18 & 50.00  \\
democratic & 16.01 & 50.00  \\
negative & 9.66 & 39.81  \\
positive & 8.45 & 60.19  \\
\hline    
\end{tabular}
\caption{Average sentence length and class distribution of style corpora.}
\label{tab:data_stats}
\end{table}


\section{Experimental Setup}
\label{sec:experiments}

In what follows, we describe our experimental settings, including baselines used, hyperparameter settings, datasets, and evaluation setups. 

\paragraph{Baseline.} We compare our model against the ``cross-aligned'' auto-encoder \cite{shen2017style}, which uses style-specific decoders to align the style of generated sentences to the actual distribution of the style. We used the off-the-shelf sentiment model released by \citet{shen2017style} for the sentiment experiments. We also separately train this model for the gender and political slant using hyper-parameters detailed below.\footnote{In addition, we compared our model with the current state-of-the-art approach introduced  by \citet{hu2017toward}; \citet{shen2017style} use this method as baseline, obtaining comparable results. We reproduced the results reported in \cite{hu2017toward} using their tasks and data. However, the same model trained on our political slant datasets (described in \Sref{sec:tasks}), obtained an almost random accuracy of 50.98\% in style transfer. We thus omit these results.} 

\paragraph{Translation data.} 
We trained an English--French neural machine translation system and a French--English back-translation system. We used data from Workshop in Statistical Machine Translation 2015 (WMT15) \cite{bojar-EtAl:2015:WMT} to train our translation models. We used the French--English data from the Europarl v7 corpus, the news commentary v10 corpus and the common crawl corpus from WMT15. Data were tokenized using the Moses tokenizer \cite{koehn2007moses}. Approximately 5.4M English--French parallel sentences were used for training. A vocabulary size of 100K was used to train the translation systems.

\paragraph{Hyperparameter settings.} In all the experiments, the generator and the encoders are  a two-layer LSTM with an input size of 300 and the hidden dimension of 500. The encoders are bidirectional. The generator samples a sentence of maximum length 50. All the generators use global attention vectors of size 500. 
The CNN classifier is trained with 100 filters of size 5, with max-pooling. The input to CNN is of size 302: the 300-dimensional word embedding plus two bits for membership of the word in our style lexicons, as described in \Sref{sec:classifier}. Balancing parameter $\lambda_{c}$ is set to 15. For sentiment task, we have used settings provided in \cite{shen2017style}.

\section{Results}
\label{sec:results}
We evaluate our approach along three dimensions.  
(1) Style transfer accuracy, measuring the proportion of our models' outputs that generate sentences of the desired style. The style transfer accuracy is performed 
using classifiers trained on held-out train data that were not used in training the style transfer models. 
(2) Preservation of meaning. (3) Fluency, measuring the readability and the naturalness of the generated sentences. 
We conducted human evaluations for the latter two.

In what follows, we first present the quality of our neural machine translation systems, then we present the evaluation setups, and then present the results of our experiments. 
\paragraph{Translation quality.}
The BLEU scores achieved for English--French MT system is 32.52 and for French--English MT system is 31.11; these are strong translation systems. We deliberately chose a European language close to English for which massive amounts of parallel data are available and translation quality is high, to concentrate on the style generation, rather than improving a translation system. \footnote{Alternatively, we could use a pivot language that is typologically more distant from English, e.g., Chinese. In this case we hypothesize that stylistic traits would be even less preserved in translation, but the quality of back-translated sentences would be worse. We have not yet investigated how the accuracy of the translation model, nor the language of translation affects our models.} 
\ignore{
\begin{table}[h!]
\begin{center}
\begin{tabular}{ c | c }
\hline
Model & BLEU \\
\hline
English--French & 32.52 \\
French--English & 31.11 \\
\hline
\end{tabular}
\end{center}
\caption{BLEU scores for translation systems.}
\label{mt_bleu}
\end{table}
}

\subsection{Style Transfer Accuracy}

We measure the accuracy of style transfer for the generated sentences using a pre-trained style classifier (\Sref{sec:classifier}). The classifier is trained on data that is not used for training our style transfer generative models (as described in \Sref{sec:tasks}). The classifier has an accuracy of 82\% for the gender-annotated corpus, 92\% accuracy for the political slant dataset and 93.23\% accuracy for the sentiment dataset.

We transfer the style of test sentences and then test the classification accuracy of the generated sentences for the opposite label. For example, if we want to transfer the style of male Yelp reviews to female, then we use the fixed common encoder of the back-translation model to encode the test male sentences and then we use the female generative model to generate the female-styled reviews. We then test these generated sentences for the \textit{female} label using the gender classifier.
\begin{table}[h!]
\begin{center}
\begin{tabular}{ c | c | c }
\hline
Experiment & CAE & BST \\
\hline
Gender & \textbf{60.40} & 57.04 \\
Political slant & 75.82 & \textbf{88.01} \\
Sentiment & 80.43 & \textbf{87.22} \\
\hline
\end{tabular}
\end{center}
\caption{Accuracy of the style transfer in generated sentences.}
\label{accuracy_result}
\end{table}

In Table \ref{accuracy_result}, we detail the accuracy of each classifier on generated style-transfered sentences.\footnote{In each experiment, we report aggregated results across directions of style transfer; same results broke-down to style categories are listed in the Supplementary Material.} 
We denote the \possessivecite{shen2017style} \textbf{C}ross-aligned \textbf{A}uto-\textbf{E}ncoder model as CAE and our model as \textbf{B}ack-translation for \textbf{S}tyle \textbf{T}ransfer (BST). 

On two out of three tasks our model substantially outperforms the baseline, by up to 12\% in political slant transfer, and by up to 7\% in sentiment modification.

\subsection{Preservation of Meaning} 
Although we attempted to use automatics measures to evaluate how well meaning is preserved in our transformations; measures such as BLEU \cite{papineni2002bleu} and Meteor \cite{denkowski2011meteor}, or even cosine similarity between distributed representations of sentences do not capture this distance well. 

Meaning preservation in style transfer is not trivial to define as literal meaning is likely to change when style transfer occurs.  For example ``My girlfriend loved the desserts'' vs ``My partner liked the desserts''.  Thus we must relax the condition of literal meaning to \emph{intent} or \emph{affect} of the utterance within the context of the discourse. Thus if the intent is to criticize a restaurant's service in a review, changing ``salad'' to ``chicken'' could still have the same effect but if the intent is to order food that substitution would not be acceptable. Ideally we wish to evaluate transfer within some downstream task and ensure that the task has the same outcome even after style transfer.  
This is a hard evaluation and hence we resort to a simpler evaluation of the ``meaning'' of the sentence.

We set up a manual pairwise comparison following \citet{bennett2005large}. The test presents the original sentence and  then, in  random order, its corresponding  sentences produced by the baseline and our models.   For the gender style transfer we asked ``Which transferred sentence maintains the same sentiment of the source sentence in the same semantic context (i.e. you can ignore if food items are changed)''.  For the task of changing the political slant, we asked ``Which transferred sentence maintains the same semantic intent of the source sentence while changing the political position''. For the task of sentiment transfer we have followed the annotation instruction in \cite{shen2017style} and asked ``Which transferred sentence is semantically equivalent to the source sentence with an opposite sentiment''

We then count the preferences of the eleven participants, measuring the relative acceptance of the generated sentences.\footnote{None of the human judges are authors of this paper}   
A third option ``=" was given to participants to mark no preference for either of the generated sentence. 
The ``no preference'' option includes choices both are equally bad and both are equally good. 
We conducted three tests one for each type of experiment - gender, political slant and sentiment. 
We also divided our annotation set into short (\#tokens $\le$ 15) and long (15 $<$ \#tokens $\le$ 30) sentences for the gender and the political slant experiment. In each set we had 20 random samples for each type of style transfer.  In total we had 100 sentences to be annotated. Note that we did not ask about appropriateness of the style transfer in this test, or fluency of outputs, only about meaning preservation.  

\begin{table}[t]
\begin{center}
\begin{tabular}{ c | c | c | c }
\hline
Experiment & CAE & No Pref. & BST \\
\hline
Gender & 15.23 & 41.36 & \textbf{43.41} \\
Political slant & 14.55 & \textbf{45.90} & 39.55 \\
Sentiment & 35.91 & \textbf{40.91} & 23.18 \\
\hline
\end{tabular}
\end{center}
\caption{Human preference for meaning preservation in percentages.}
\label{tab:human-meaning-results}
\end{table}

The results of human evaluation are presented in Table \ref{tab:human-meaning-results}. Although a no-preference option was chosen often---showing that state-of-the-art systems are still not on par with human expectations---the BST models outperform the baselines in the gender and the political slant transfer tasks. 

Crucially, the BST models significantly outperform the CAE models when transferring style in longer and harder sentences. Annotators preferred the CAE model only for 12.5\% of the long sentences, compared to 47.27\% preference for the BST model. 



\subsection{Fluency} 
Finally, we evaluate the fluency of the generated sentences. Fluency was rated from 1 (unreadable) to 4 (perfect) as is described in \cite{shen2017style}. We randomly selected 60 sentences each generated by the baseline and the BST model. 

The results shown in Table \ref{fluency_result} are averaged scores for each model. 
\begin{table}[h!]
\begin{center}
\begin{tabular}{ c | c | c }
\hline
Experiment & CAE & BST \\
\hline
Gender & 2.42 & \textbf{2.81} \\
Political slant & 2.79 & \textbf{2.87} \\
Sentiment & 3.09 & \textbf{3.18} \\
Overall & 2.70 & \textbf{2.91} \\
Overall Short & 3.05 & \textbf{3.11} \\
Overall Long & 2.18 & \textbf{2.62} \\
\hline
\end{tabular}
\end{center}
\caption{Fluency of the generated sentences.}
\label{fluency_result}
\end{table}

BST outperforms the baseline overall. It is interesting to note that BST generates significantly more fluent longer sentences than the baseline model. Since the average length of sentences was higher for the gender experiment, BST notably outperformed the baseline in this task, relatively to the sentiment task where the sentences are shorter.
Examples of the original and style-transfered sentences generated by the baseline and our model are shown in the Supplementary Material. 

\subsection{Discussion} The loss function of the generators given in Eq. \ref{recon_loss} includes two competing terms, one to improve meaning preservation and the other to improve the style transfer accuracy. In the task of sentiment modification, the BST model preserved meaning worse than the baseline, on the expense of being better at style transfer. We note, however, that the sentiment modification task is not particularly well-suited for evaluating style transfer: it is particularly hard (if not impossible) to disentangle the sentiment of a sentence from its propositional content, and to modify sentiment while preserving meaning or intent. On the other hand, the style-transfer accuracy for gender is lower for BST model but the preservation of meaning is much better for the BST model, compared to CAE model and to "No preference" option. This means that the BST model does better job at closely representing the input sentence while taking a mild hit in the style transfer accuracy. 

\section{Related Work}

Style transfer with non-parallel text corpus has become an active research area due to the recent advances in text generation tasks. \citet{hu2017toward} use variational auto-encoders with a discriminator to generate sentences with controllable attributes. The method learns a disentangled latent representation and generates a sentence from it using a code. This paper mainly focuses on sentiment and tense for style transfer attributes. It evaluates the transfer strength of the generated sentences but does not evaluate the extent of preservation of meaning in the generated sentences. In our work, we show a qualitative evaluation of meaning preservation.

\citet{shen2017style} first present a theoretical analysis of style transfer in text using non-parallel corpus. The paper then proposes a novel cross-alignment auto-encoders with discriminators architecture to generate sentences. It mainly focuses on sentiment and word decipherment for style transfer experiments. 

\citet{FuTan} explore two models for style transfer. 
The first approach uses multiple decoders for each type of style. In the second approach, style embeddings are used to augment the encoded representations, so that only one decoder needs to be learned to generate outputs in different styles. Style transfer is evaluated on scientific paper titles and newspaper tiles, and sentiment in reviews. This method is  different from ours in that we use machine translation to create a strong latent state from which multiple decoders can be trained for each style. 
We also propose a different human evaluation scheme.

\citet{HeHe} first extract words or phrases associated with the original style of the sentence, delete them from the original sentence and then replace them with new phrases associated with the target style. 
They then use a neural model to fluently combine these into a final output. 
\citet{2017arXiv170604223J} learn a representation which is style-agnostic, using adversarial training of the auto-encoder.

Our work is also closely-related to a problem of paraphrase generation \cite{madnani2010generating,mirella}, including methods relying on (phrase-based) back-translation \cite{ganitkevitch2011learning,ganitkevitch2014multilingual}.
More recently, \citet{mallinson2017paraphrasing} and \citet{WietingMG17} showed how neural back-translation can be used to generate paraphrases. 
An additional related line of research is machine translation with non-parallel data. \citet{guillaume} and \citet{artetxe2017unsupervised} have proposed sophisticated methods for unsupervised machine translation. These methods could in principle be used for style transfer as well. 

\section{Conclusion}

We propose a novel approach to the task of style transfer with non-parallel text.\footnote{All the code and data used in the experiments will be released to facilitate reproducibility at https://github.com/shrimai/Style-Transfer-Through-Back-Translation} 
We learn a latent content representation using machine translation techniques; this aids grounding the meaning of the sentences, as well as weakening the style attributes.
We apply this technique to three different style transfer tasks. In transfer of political slant and sentiment we outperform an off-the-shelf state-of-the-art baseline using a cross-aligned autoencoder. The political slant task is a novel task that we introduce. 
Our model also outperforms the baseline in all the experiments of fluency, and
in the experiments for meaning preservation in generated sentences of gender and political slant. 
Yet, we acknowledge that the generated sentences do not always adequately preserve meaning.

This technique is suitable not just for style transfer, but for enforcing style, and removing style too. In future work we intend to apply this technique to \emph{debiasing} sentences and \emph{anonymization} of author traits such as gender and age.

In the future work, we will also explore whether an enhanced back-translation by pivoting through several languages will learn better grounded latent meaning representations.
In particular, it would be interesting to back-translate through multiple target languages with a single source language \cite{johnson2016google}. 

Measuring the separation of style from content is hard, even for humans.  It depends on the task and the context of the utterance within its discourse. Ultimately we must evaluate our style transfer within some down-stream task where our style transfer has its intended use but we achieve the same task completion criteria.  

\section*{Acknowledgments}
This work was funded by a fellowship from Robert Bosch, and in part by the National
Science Foundation through award IIS-1526745. We would like to thank Sravana Reddy for sharing the Yelp corpus used in gender transfer experiments, Zhiting Hu for providing an implementation of a VAE-based baseline, and  the 11 CMU graduate students who helped with annotation and manual evaluations. We are also grateful to the anonymous reviewers for their constructive feedback, and to Dan Jurafsky, David Jurgens, Vinod Prabhakaran, and Rob Voigt for valuable discussions at earlier stages of this work.

\bibliography{acl2018}
\bibliographystyle{acl_natbib}

\appendix
\newpage
\section{Supplementary Material}
\label{sec:supplemental}

In Tables~\ref{gender_result},~\ref{fb_result}, and~\ref{sentiment_result} we present the style transfer accuracy results broken-down to style categories.
We denote the \textbf{C}ross-aligned \textbf{A}uto-\textbf{E}ncoder model as CAE and our model as \textbf{B}ack-translation for \textbf{S}tyle \textbf{T}ransfer (BST). 

\begin{table}[h!]
\begin{center}
\begin{tabular}{ c | c | c }
\hline
Model & Style transfer & Acc \\
\hline
CAE & male $\rightarrow$ female & \textbf{64.75} \\
BST & male $\rightarrow$ female & 54.59 \\
\hline
CAE & female $\rightarrow$ male & 56.05 \\
BST & female $\rightarrow$ male & \textbf{59.49} \\
\hline
\end{tabular}
\end{center}
\caption{Gender transfer accuracy.}
\label{gender_result}
\end{table}

\begin{table}[h!]
\begin{center}
\begin{tabular}{ c | c | c }
\hline
Model & Style transfer & Acc \\
\hline
CAE & republican $\rightarrow$ democratic & 65.44 \\
BST & republican $\rightarrow$ democratic & \textbf{80.55} \\\hline 
CAE & democratic $\rightarrow$ republican & 86.20 \\
BST & democratic $\rightarrow$ republican & \textbf{95.47} \\
\hline
\end{tabular}
\end{center}
\caption{Political slant transfer accuracy.}
\label{fb_result}
\end{table}

\begin{table}[h!]
\begin{center}
\begin{tabular}{ c | c | c }
\hline
Model & Style transfer & Acc \\
\hline
CAE & negative $\rightarrow$ positive & 81.63 \\
BST & negative $\rightarrow$ positive & \textbf{95.68} \\\hline 
CAE & positive $\rightarrow$ negative & 79.65 \\
BST & positive $\rightarrow$ negative & \textbf{81.65} \\
\hline
\end{tabular}
\end{center}
\caption{Sentiment modification accuracy.}
\label{sentiment_result}
\end{table}

In Table \ref{gender_old}, we detail the accuracy of the gender classifier on generated style-transfered sentences by an auto-encoder; Table \ref{fb_old} shows the accuracy of transfer of political slant.
The experiments are setup as described in Section 5.1.
We denote the Auto-Encoder as (AE) and our model as Back-translation for Style Transfer (BST).
\begin{table}[h!]
\begin{center}
\begin{tabular}{ c | c | c }
\hline
Model & Style transfer & Acc \\
\hline
AE & male $\rightarrow$ female & 41.48 \\
BST & male $\rightarrow$ female & \textbf{54.59} \\
\hline
AE & female $\rightarrow$ male & 41.88 \\
BST & female $\rightarrow$ male & \textbf{59.49} \\
\hline
\end{tabular}
\end{center}
\caption{Gender transfer accuracy for Auto-encoder.}
\label{gender_old}
\end{table}

\begin{table}[h!]
\begin{center}
\begin{tabular}{ c | c | c }
\hline
Model & Style transfer & Acc \\
\hline
AE & republican $\rightarrow$ democratic & 60.76 \\
BST & republican $\rightarrow$ democratic & \textbf{80.55} \\\hline 
AE & democratic $\rightarrow$ republican & 64.05 \\
BST & democratic $\rightarrow$ republican & \textbf{95.47} \\
\hline
\end{tabular}
\end{center}
\caption{Political slant transfer accuracy for Auto-encoder.}
\label{fb_old}
\end{table}

To evaluate the preservation of meaning by the Auto-Encoder, the experiments were setup as described in Section 5.2. We conducted four tests, each of 20 random samples for each type of style transfer.  Note that we did not ask about appropriateness of the style transfer in this test, or fluency of outputs, only about meaning preservation.  We show the results of human evaluation in Table \ref{tab:human-meaning-results_old}
\begin{table}[h]
\centering
\begin{tabular}{c|c|c|c}
\hline 
Style transfer              & = & AE  & BST \\
\hline
male $\rightarrow$ female     & 43.3 & 13.4 & 43.3 \\
female $\rightarrow$ male     & 45.0 & 10.0 & 45.0 \\ 
\hline
republican $\rightarrow$ democratic & 43.3 & 3.4 & \textbf{53.3} \\
democratic $\rightarrow$ republican  & \textbf{55.00} & 11.7 & 33.3 \\ 
\hline
\end{tabular}
\caption{Human preference for meaning preservation in percentages.}
\label{tab:human-meaning-results_old}
\end{table}

Examples of the original and style-transfered sentences generated by the baseline and our model are shown in Table \ref{style_samples}
\begin{table*}[h!]
\begin{tabular}{ p{5cm} | p{5cm} | p{5cm} }
\hline
\multicolumn{1}{c|}{Input Sentence} & \multicolumn{1}{c|}{CAE}  & \multicolumn{1}{c}{BST} \\
\hline
\multicolumn{3}{c}{ male  $\rightarrow$ female}\\
\hline
\textit{my wife ordered country fried} & \textit{i got ta get the chicken breast .} & \textit{my husband ordered the chicken} \\
\excont{steak and eggs.} & \excont{} &  \excont{salad and the fries.}\\
\textit{great place to visit and maybe} & \textit{we could n't go back and i would} & \textit{great place to go back and try a}\\
\excont{find that one rare item you} & \excont{be able to get me to get me.} & \excont{lot of which you ' ve never had}\\
\excont{just have never seen or can} & & \excont{to try or could not have been}\\
\excont{not find anywhere else.} & & \excont{able to get some of the best.}\\
\textit{the place is small but cosy and} \excont{very clean.} &
\textit{the staff and the place is very} \excont{nice.} &
\textit{the place is great but very clean} \excont{and very friendly.} \\
\hline
\multicolumn{3}{c}{ female  $\rightarrow$ male}\\
\hline
\textit{save yourself the huge} \excont{headaches.} & \textit{the sauces are excellent.} & \textit{you are going to be} \excont{disappointed.} \\
\textit{would i discourage someone else} & \textit{i believe i would be back?} & \textit{i wouldn't go back!}\\
\excont{from going?} &  & \\
\textit{my husband ordered the salad} \excont{and the dressing - lrb - blue} \excont{cheese - rrb - was watered} \excont{down.} & \textit{the sauces - lrb - - rrb - - rrb -} \excont{and - rrb -.} & \textit{my wife ordered the mac-n-}\excont{cheese and the salad - lrb - \$} \excont{00 minutes - rrb - was cooked.}\\
\hline
\multicolumn{3}{c}{ republican  $\rightarrow$ democratic}\\
\hline
\textit{i will continue praying for you} & \textit{i am proud of you and your vote} & \textit{i will continue to fight for you} \\
\excont{and the decisions made by}  &  \excont{for us!} &  \excont{and the rest of our} \\
\excont{our government!} & &   \excont{democracy!}\\
\textit{tom, i wish u would bring} \excont{change.}& \textit{i agree, senator warren and} \excont{could be.} & \textit{brian, i am proud to have you} \excont{representing me.}\\
\textit{all talk and no action-why dont} \excont{you have some guts like} \excont{breitbart} & \textit{and then we will be praying for} \excont{them and i am proud of this} \excont{position and i am proud of} &
\textit{keep up and don’t know, you} \excont{have a lot of respect as} \excont{breitbart}\\
\hline
\multicolumn{3}{c}{{democratic}  $\rightarrow$ {republican}} \\
\hline
\textit{as a hoosier, i thank you, rep.} & \textit{a lot , i am proud of you $<$unk$>$.} & \textit{as a hoosier, i'm praying for}\\
\excont{visclosky.} &  &  \excont{you sir.} \\
\textit{thank you for standing up for} & \textit{do you for standing up for highly} & \textit{thanks for standing up for the}\\
\excont{justice and against bigotry-} & \excont{and in bigotry--racism, } & \excont{constitution and get rid of}\\
\excont{-racism, homophobia, sexism} & \excont{programming, cut,} & \excont{obamacare, homophobie}\\
\excont{, misogyny, religious and } & \excont{granddaughters, } & \excont{, cut, and actuality.}\\
\excont{xenophobia.} &  \excont{unprescedented and excludes.} &  \\
\textit{thank you for all you are doing} \excont{for us, attorney general} \excont{harris!} & \textit{thank you for standing up for us} \excont{and i am proud of us!} & \textit{thanks lawmaker for all you do} \excont{for us, senator scott!} \\
\hline
\multicolumn{3}{c}{{negative}  $\rightarrow$ {positive}} \\
\hline
\textit{crap fries, hard hamburger buns,} \excont{burger tasted like crap!} & \textit{good selection, fresh food, like} \excont{like like!} & \textit{empathy, the best food,} \excont{but it was very nice!} \\
\textit{the people behind the counter} \excont{were not friendly whatsoever.} & \textit{the people who the staff were} \excont{friendly.} & \textit{the people here are really good.} \\
\textit{this place is bad news!} &
\textit{this place is great!} &
\textit{this place is amazing!} \\
\hline
\multicolumn{3}{c}{{positive}  $\rightarrow$ {negative}} \\
\hline
\textit{the food is excellent and the} \excont{service is exceptional!} & \textit{the food is the food and the} \excont{service is terrible.} & \textit{the food is horrible and the} \excont{service is terrible.} \\
\textit{great as always, i love there} \excont{food.} & \textit{horrible as, i really don't eat} \excont{here.} & \textit{really disappointed, i couldn't be} \excont{back.} \\
\textit{i would recommend a visit here.} &
\textit{i would not recommend a dinner here.} & \textit{i will not recommend this place.} \\
\hline
\end{tabular}
\caption{Gender, Political slant and Sentiment style transfer examples. In addition to better preserving meaning, sentences generated by the BST model are generally grammatically better structured. 
}
\label{style_samples}
\end{table*}

\end{document}